\PassOptionsToPackage{table,xcdraw}{xcolor}

\documentclass[10pt,logo,copyright]{nvidiatechreport}

\usepackage[numbers, sort]{natbib}

\usepackage[capitalize]{cleveref}
\crefname{section}{\S}{\S\S}
\crefname{subsection}{\S}{\S\S}
\crefname{table}{\text{Tab.}}{\text{Tab.}}
\Crefname{table}{Table}{Tables}
\crefname{figure}{\text{Fig.}}{\text{Fig.}}
\Crefname{figure}{Figure}{Figures}
\crefname{equation}{\text{Eq}}{\text{Eq}}

\newcommand{\ours}{BirdDreams\xspace}

\DeclareRobustCommand{\rankfirst}[1]{\textbf{#1}}
\DeclareRobustCommand{\ranksecond}[1]{\underline{#1}}

\newcommand{\titlebreak}{\\}

\title{BirdDreams: Joint Perspective and BEV Video Generation \titlebreak for a Plausible, Reactive Driving World Model}


\author{%
Junyoung Seo\textsuperscript{2*}, Jaewoo Seo\textsuperscript{1}, Seungryong Kim\textsuperscript{2}, Laura Leal-Taixé\textsuperscript{1}, Xiuming Zhang\textsuperscript{1\textdagger}, Seung Wook Kim\textsuperscript{2*\textdagger}\\
\small \textsuperscript{1}NVIDIA \quad \textsuperscript{2}KAIST \\
\small \href{https://research.nvidia.com/labs/sil/projects/birddreams/}{https://research.nvidia.com/labs/sil/projects/birddreams/}
}

\begin{document}
\maketitle
\begingroup
\renewcommand{\thefootnote}{\fnsymbol{footnote}}
\footnotetext[1]{Work done while at NVIDIA.}
\footnotetext[2]{Corresponding authors.}
\endgroup

\begin{abstract}
Driving world models aim to simulate how a scene unfolds under different ego actions. This requires plausible motion of surrounding vehicles as well as precise ego-trajectory control. Learning vehicle motion from perspective video alone is challenging because motion is entangled with appearance and vehicles can be occluded. Prior work therefore conditions video generation on layouts that specify vehicle motion, providing control over surrounding vehicles but requiring their future trajectories as inputs. We introduce \ours, an autoregressive driving world model that jointly generates perspective video and a schematic bird's-eye-view (BEV) video, making scene layout part of the predicted future. The BEV, rendered from scene annotations for training, depicts roads and vehicles as simple shapes at a fixed metric scale, providing a top-down view of their positions and motion with reduced occlusion between vehicles. Perspective video, by contrast, retains scene details that the BEV omits, such as traffic signals and object appearance. Joint attention between two video diffusion transformers, one per stream, lets each stream inform the other during generation. The predicted BEV also provides a geometric interface for ego control: we translate and rotate its latest frame along an input ego trajectory and use the resulting frames as a soft condition that shows how the scene would appear to move if only the ego vehicle moved. At inference, \ours generates both videos from a single camera image without requiring an HD map or vehicle-state inputs. On PhysicalAI-AV and nuScenes, BirdDreams follows input ego trajectories more faithfully than prior driving world models and generates more plausible ego motion without ego-action conditioning. Human evaluation shows that BirdDreams outperforms a perspective-only baseline in generating realistic surrounding-vehicle behavior.

\end{abstract}

\abscontent

\suppressfloats[t]
\begin{figure}[t]
  \centering
  \includegraphics[width=\linewidth]{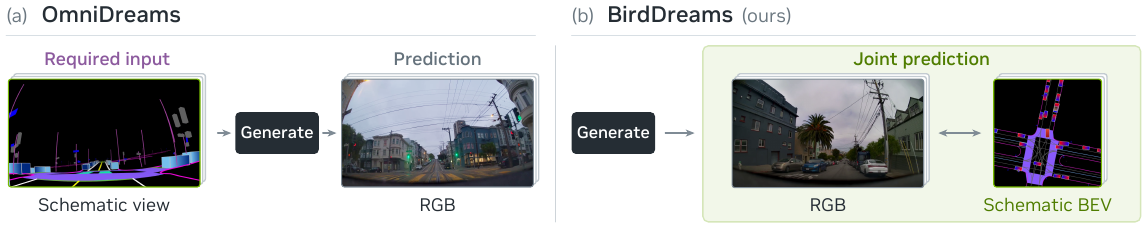}
  \caption{\textbf{Motivation: scene structure as a prediction target.} (a)~OmniDreams conditions RGB video generation on schematic views rendered from external road geometry and future vehicle states. (b)~\ours jointly generates perspective and schematic BEV videos, predicting scene structure and surrounding-vehicle motion without HD-map or layout inputs at inference. Schematic BEV targets supervise vehicle motion during training, while the predicted BEV provides a geometric interface for ego-trajectory control.}
  \label{fig:teaser}
\end{figure}

\section{Introduction}
\label{sec:intro}

Simulation allows autonomous driving systems to be tested in rare, dangerous scenarios that are difficult to reproduce safely on real roads. Driving world models learn from real driving videos to synthesize realistic future observations conditioned on ego actions~\cite{gao2024vista,russell2025gaia}. From the same initial observation, they can simulate how a scene unfolds under different ego maneuvers. Such simulation requires the model to predict plausible responses of surrounding vehicles to the ego maneuver while generating ego motion that follows the input trajectory. At an intersection, for example, an ego vehicle proceeding instead of yielding may cause a crossing vehicle to slow down or change course.

Recent driving world models such as OmniDreams~\cite{omnidreams} condition video generation on external scene layouts that specify road geometry and surrounding-vehicle motion. Replaying these layouts from driving logs fixes surrounding vehicles to their recorded trajectories, so they cannot react to changes in the ego maneuver. A separate traffic model can generate new trajectories, but leaves interaction prediction outside the video world model~\cite{yang2025drivearena}. Without external scene-layout inputs, the world model must learn surrounding-vehicle motion from perspective video. This is difficult because motion is entangled with texture and lighting, occlusion often hides vehicles, and the same pixel displacement can correspond to different speeds depending on a vehicle's distance from the camera. In our OmniDreams baseline trained without surrounding-vehicle trajectories as inputs, vehicles can disappear or make implausible maneuvers at complex intersections (\cref{fig:counterfactual-baselines}).

We introduce \ours, an autoregressive driving world model that jointly generates video from the ego vehicle's perspective and schematic bird's-eye-view (BEV) video of the same scene. Scene layout thus becomes part of the generated future rather than an external input (\cref{fig:teaser}). The schematic BEV depicts road elements and vehicles as simple shapes with semantic colors. This simplified prediction task is intended to help the model focus on learning scene dynamics rather than reproducing detailed appearance. In addition, the BEV's fixed metric scale makes a given pixel displacement correspond to the same distance anywhere in the view, unlike in perspective video. Its top-down view also reduces occlusion between vehicles. At a complex intersection, for example, crossing vehicles that overlap in the perspective view can appear separately in the BEV, making their paths relative to the road easier to follow. Photorealistic perspective video, by contrast, contains visual information that the schematic representation omits, such as road-surface details and distant objects beyond the BEV's range. Joint prediction combines the BEV's representation of vehicle motion with this visual information from perspective video.

To let each stream shape the other during generation, \ours couples two video diffusion-transformer branches, one for perspective video and one for schematic BEV video, through joint attention throughout denoising. The perspective branch can thus use vehicle motion from the BEV, and the BEV branch can use visual features from the perspective video. We train both branches to predict synchronized perspective and BEV videos, with the BEV rendered from scene annotations. At inference, the model generates both videos from a single image, without requiring an HD map or vehicle-state inputs.

Accurately following an ego trajectory specified in meters also remains challenging for video world models. Even with trajectory conditioning, subtle steering maneuvers may not be faithfully reproduced in the generated ego motion. We use the predicted BEV as an action interface for video generation. In an ego-centered, orthographic BEV at a fixed metric scale, a change in ego position shifts the scene by a uniform number of pixels, and a change in heading rotates it about the ego vehicle. Our BEV-aware action conditioning uses these transformations to reproject the latest predicted BEV frame into the ego coordinate frame at each pose along the input trajectory. The resulting frames show how the scene would appear to move if the surroundings stayed fixed. They serve as a soft condition for the BEV branch, and the perspective branch attends to the conditioned BEV features through joint attention. The model predicts how surrounding vehicles move and interact in both videos. Without ego-action conditioning, ego motion is also predicted as part of the rollout.

We evaluate \ours on PhysicalAI-AV~\cite{physicalaiav} and nuScenes~\cite{nuscenes}. Compared with open driving world models and OmniDreams-based baselines that generate perspective video alone, it precisely follows input ego trajectories specified in meters and, without them, generates more plausible ego motion. When the ego vehicle departs from the logged trajectory, surrounding vehicles in its generated videos can respond to the new maneuver, as when a following vehicle slows down after the ego vehicle cuts into its lane. Human evaluation further demonstrates that \ours generates more realistic surrounding-vehicle motion and behavior than an OmniDreams baseline under the same ego maneuvers.

\section{Related Work}
\label{sec:related}

\noindent\textbf{Video world models.}
Video world models formulate interactive simulation as predicting future visual observations under actions, as demonstrated by GameNGen~\cite{valevski2025gamengen} and Genie~\cite{bruce2024genie}.
Oasis~\cite{decart2024oasis} and MineWorld~\cite{guo2025mineworld} generate interactive gameplay; Genie~3~\cite{genie3} supports diverse interactive worlds.
Matrix-Game~2.0~\cite{he2025matrixgame2}, WorldPlay~\cite{sun2025worldplay}, and LingBot-World~\cite{robbyant2026lingbot} target real-time interaction and long-horizon generation.
Sora~\cite{openai2024sora}, Movie Gen~\cite{polyak2024moviegen}, Wan~\cite{wan2025wan}, and Veo~3~\cite{collins2025veo3} generate general-purpose videos from text or images.
Large-scale pretraining in Cosmos-Predict2.5, and Cosmos~3~\cite{ali2025world,nvidia2026cosmos3} provides foundations for adapting video world models to robotics and driving.
Policy learning draws on both the simulated experience produced by world models and the representations learned during video pretraining.
DIAMOND~\cite{alonso2024diamond} trains agents inside a diffusion world model.
Dreamer~4~\cite{hafner2025training} learns policies through reinforcement learning on imagined rollouts of a world model trained on offline data.
World-action models adapt pretrained video backbones to jointly predict video and actions, as in DreamZero~\cite{ye2026world} for robotics and DriveVA~\cite{liu2026driveva} for driving.
OmniDreams~\cite{omnidreams} also transfers its video backbone to ego-trajectory prediction through post-training with a trajectory head.

\noindent\textbf{Driving world models.}
Driving world models use conditional video generation to simulate common and rare driving scenarios~\cite{russell2025gaia}.
DriveGAN~\cite{kim2021drivegan} was an early generative driving simulator learned from real-world video, supporting ego-motion and scene-content control.
GAIA-1~\cite{hu2023gaia1} and GAIA-2~\cite{russell2025gaia} condition on actions and scenes; Drive-WM~\cite{wang2024drivewm} uses multiview prediction for planning.
DriveDreamer~\cite{wang2024drivedreamer} and MagicDrive~\cite{gao2024magicdrive} provide geometry-controlled scene generation.
Cosmos-Drive-Dreams~\cite{ren2025cosmosdrivedreams} generates synthetic perception and policy-training data.
The Waymo World Model~\cite{waymo2026worldmodel} adapts Genie~3~\cite{genie3} for camera/lidar generation with action, layout, and language control.
In Panacea~\cite{wen2024panacea} and OmniDreams~\cite{omnidreams}, conditioning on road and vehicle layouts provides direct control over scene geometry and surrounding-vehicle motion.
Logged layouts are useful for reconstructing recorded scenes, but replaying them under a different ego maneuver does not account for how surrounding vehicles might brake or yield in response.
DriveArena~\cite{yang2025drivearena} updates the layouts with an external traffic model, supporting reactive simulation while keeping interaction prediction outside the video generator.
GenAD's action-conditioned extension~\cite{yang2024genad} and Vista~\cite{gao2024vista} instead generate future video under ego actions without future surrounding-vehicle layouts.
GAIA-4~\cite{wayve2026gaia4} supports closed-loop simulation grounded in recorded driving scenes, with optional reactive agents that respond to the ego vehicle while the rest of the scene follows the recording.
Concurrent work CausalDrive~\cite{yan2026causaldrive} focuses on how surrounding vehicles respond to counterfactual ego actions in generated video.
\ours shares this goal and makes scene layout part of the prediction, learning from synchronized perspective and schematic BEV videos as joint prediction targets.

\noindent\textbf{Structured representations for video generation.}
Physical errors can remain in visually realistic videos~\cite{bansal2026videophy}, motivating methods that model scene geometry and motion alongside appearance.
In driving, BEVWorld~\cite{zhang2024bevworld} forecasts learned scene-level BEV latents and decodes sensor observations, while GenieDrive~\cite{yang2026geniedrive} predicts future occupancy before generating video. Joint appearance–structure prediction takes several forms: GEM~\cite{gem} jointly denoises RGB and depth, UniFuture~\cite{unifuture} exchanges features between video denoising and depth prediction, and OmniNWM~\cite{li2026omninwm} jointly generates panoramic RGB, semantics, and depth, then lifts these outputs into 3D occupancy.
MMPhysVideo~\cite{lin2026mmphysvideo} jointly generates RGB and camera-aligned perceptual cues before distilling an RGB-only model.
We share this motivation for joint structural supervision, but predict an orthographic schematic BEV video rather than camera-aligned perceptual cues.
Starting from a pretrained video backbone, \ours couples perspective video with an orthographic schematic BEV stream through joint attention. The BEV makes road-relative vehicle motion explicit at a fixed metric scale without perspective foreshortening, while perspective video retains scene details omitted by the schematic representation.
Simple shapes remove texture and lighting variation; explicit headings and metric spacing expose relative motion and lane relationships, while the top-down view reduces occlusion.
This simpler target is intended to ease learning vehicle dynamics, while joint attention incorporates RGB cues such as traffic-light states.
We retain the BEV during rollout and reproject its latest prediction along the ego trajectory for action conditioning.

\section{Method}
\label{sec:method}

\begin{figure}[!t]
  \centering
  \includegraphics[width=\linewidth]{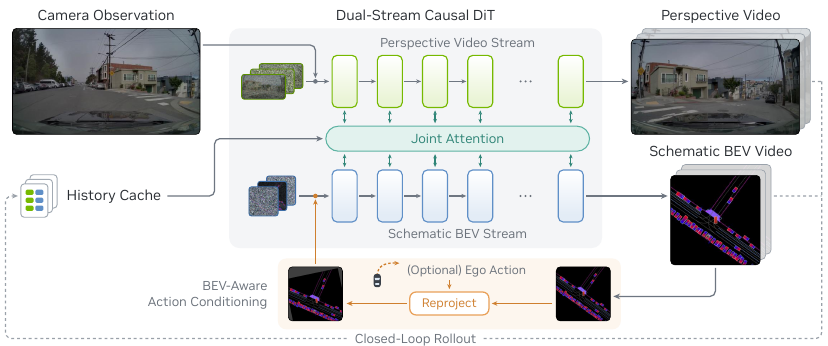}
  \caption{\textbf{Overview of \ours.} Two causal video diffusion transformers jointly generate perspective and schematic bird's-eye-view (BEV) videos from an initial camera image, with joint attention over tokens from both streams. Noise levels are sampled independently across streams during training. For ego-action conditioning, the latest predicted BEV frame is reprojected along the input ego trajectory to form a soft condition for the BEV branch. Joint attention conveys this condition to the perspective branch, and cached history supports generation of successive chunks. The VAE encoder and decoder are omitted for clarity.}
  \label{fig:method_overview}
\end{figure}

\subsection{Motivation and Overview}
\label{subsec:method_formulation}

A driving world model must predict how a scene evolves under ego actions, including the responses of surrounding vehicles. We supervise these dynamics through two complementary videos of the same scene. Photorealistic perspective video preserves the scene's appearance and visual cues such as traffic-light states. An ego-centered, orthographic schematic BEV video makes road geometry, vehicle positions, and headings explicit at a fixed metric scale. By removing texture and lighting variation from the prediction target, the schematic BEV is intended to focus learning on road-relative vehicle motion. We jointly predict both videos so that the perspective branch can use the BEV's explicit representation of scene dynamics, with the aim of generating more plausible vehicle behavior. We render the BEV training videos from the map and 3D object annotations provided with the driving datasets (\cref{subsec:method_data}).

We build on the pretrained causal video diffusion transformer of vanilla OmniDreams~\cite{omnidreams}. Given an initial perspective RGB image $I_0$, \ours jointly generates a perspective video $I_{1:T}$ and a schematic BEV video $B_{0:T}$. Optional ego actions specify a trajectory $\mathcal{A}_{0:T}=\{(p_t,\theta_t)\}_{t=0}^{T}$, where $p_t\in\mathbb{R}^2$ denotes the planar position in meters and $\theta_t$ the heading, expressed in a common coordinate frame. We model the joint conditional distribution
\begin{equation}
\label{eq:problem_formulation}
p_\phi\!\left(I_{1:T},B_{0:T}\mid I_0,c\right),
\qquad c\in\{\mathcal{A}_{0:T},\varnothing\}.
\end{equation}
With $c=\mathcal{A}_{0:T}$, the trajectory specifies the desired ego motion. With $c=\varnothing$, ego motion is predicted as part of the video generation. In either setting, surrounding vehicles' motion and responses are predicted through both video outputs. The initial BEV frame $B_0$ is generated from $I_0$, and no BEV observation, HD map, or vehicle-state inputs are required at inference.

\subsection{\ours}
\label{subsec:method_model}

\subsubsection{Causal Mixture-of-Transformers Architecture}
\label{subsec:method_architecture}

We couple perspective and BEV video generation throughout denoising so that scene dynamics learned in the BEV's metric space can guide perspective video generation. The BEV branch, in turn, can use scene details from perspective video that are absent from its schematic representation. At every transformer block, joint attention lets each branch's queries attend to keys and values from both streams. The branches retain separate attention projections, feed-forward layers, and output heads to specialize in their respective representations.

The perspective branch inherits the pretrained OmniDreams backbone, and the BEV branch is initialized with a copy of its weights. We use the pretrained video VAE from Cosmos-Predict2.5~\cite{ali2025world}, keeping its weights frozen. It encodes each stream into latents, which are partitioned into chunks of three latent frames. Attention is bidirectional within each chunk and causal across chunks, allowing the current perspective and BEV predictions to exchange information while drawing on earlier scene history.

\subsubsection{Diffusion Forcing with Per-Stream Noise Sampling}
\label{subsec:method_df}

Following Diffusion Forcing (DF)~\cite{diffusionforcing}, we train on chunks with independently sampled noise levels. We also sample the perspective and BEV noise levels independently within each chunk. Training includes pairs in which either stream is relatively clean while the other is noisy. Relatively clean BEV latents can provide road geometry and vehicle positions to the perspective branch, while relatively clean perspective latents can provide visual features to the BEV branch. When both streams are noisy, the two branches exchange information through joint attention to denoise the perspective and BEV latents together.

Let $z_k^s$ be the clean VAE latents for stream $s\in\{\mathrm{persp},\mathrm{bev}\}$ in chunk $k$. We corrupt them using the rectified-flow interpolation~\cite{rectifiedflow}:
\begin{equation}
\label{eq:flow_interpolation}
x_k^s = (1-\sigma_k^s)z_k^s + \sigma_k^s\epsilon_k^s,
\qquad \epsilon_k^s\sim\mathcal{N}(0,I),
\end{equation}
where $\sigma_k^s\in[0,1]$ is shared by the three latent frames within a chunk but sampled independently across chunks and streams. Each branch receives an embedding of its own noise level.

The model predicts the flow velocity $\hat v_k^s$ with target $\epsilon_k^s-z_k^s$. For a training sequence with $K$ chunks, the joint flow-matching loss is
\begin{equation}
\label{eq:training_objective}
\mathcal{L}_{\mathrm{flow}}
= \mathbb{E}_{z,\sigma,\epsilon}\!\left[
\tfrac{1}{2K}\sum_{s\in\{\mathrm{persp},\mathrm{bev}\}}\sum_{k=1}^{K}
\tfrac{1}{N_s}\left\|\hat v_k^s-(\epsilon_k^s-z_k^s)\right\|_2^2
\right],
\end{equation}
where $N_s$ is the number of latent elements in a chunk of stream $s$. The expectation is over paired training latents $z=\{z_k^s\}_{k,s}$, noise levels $\sigma=\{\sigma_k^s\}_{k,s}$ drawn from the training noise schedule, and independent Gaussian noise $\epsilon=\{\epsilon_k^s\}_{k,s}$. Within each sample, we average the squared error over latent elements and chunks within each stream, then weight the two streams equally. This prevents the larger perspective frame from receiving more weight simply because it contains more latent elements.

\begin{figure}[t]
  \centering
  \includegraphics[width=\linewidth]{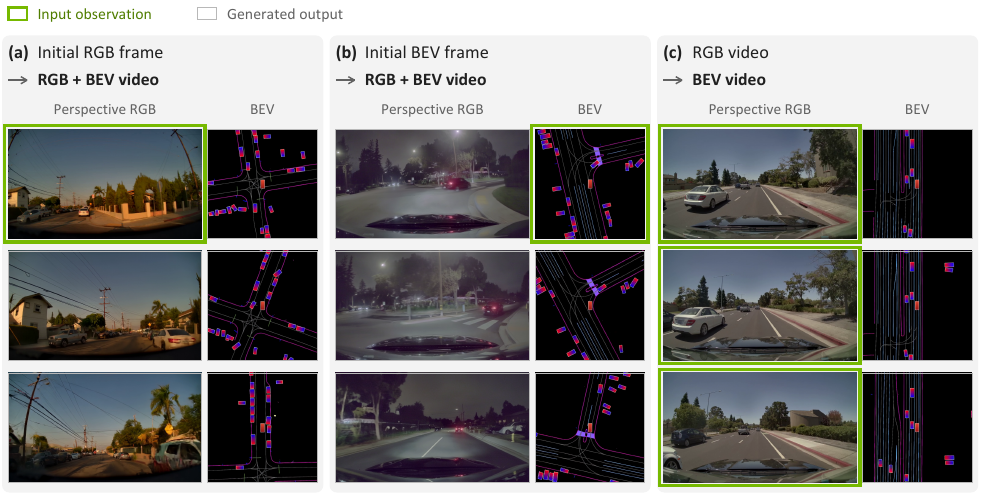}
  \caption{\textbf{Flexible conditioning.} We show example rollouts from an initial RGB frame~(a) and an initial BEV frame~(b), generating both perspective and BEV videos in each case. In~(c), we generate BEV video conditioned on a full RGB video.}
  \label{fig:flexible-conditioning}
\end{figure}

\subsubsection{BEV-Aware Action Conditioning}
\label{subsec:method_action}

One goal of joint perspective--BEV video modeling is to control ego motion in perspective video through the BEV's metric geometry. The BEV's fixed metric scale and orthographic projection let us translate and rotate a previously predicted frame according to the input ego trajectory. This reprojection accounts for ego-induced apparent motion while treating the reference scene as stationary. We use the reprojected BEV as a soft condition and train the model to predict the actual future scene, including surrounding-vehicle motion and newly visible content. This follows the warped-render conditioning idea of Gen3C~\cite{gen3c}, using planar BEV geometry without depth estimation.

The initial BEV $B_0$ generated from $I_0$ serves as the reference for the first chunk. For later chunks, we use the last BEV frame $B_r$ of the most recently generated chunk as the reference. Let $R(\theta)$ be the two-dimensional rotation matrix and $\Pi$ the mapping from ego coordinates in meters to BEV pixels, with the ego centered and its heading pointing upward. We reproject $B_r$ to the ego pose $(p_t,\theta_t)$, as illustrated in \cref{fig:method_overview}, by mapping each target pixel $u$ back to the reference frame:
\begin{equation}
\label{eq:bev_reprojection}
\widetilde B_t(u) = B_r\!\left(
\Pi\!\left[
R(\theta_r)^{\mathsf T}
\left(R(\theta_t)\Pi^{-1}(u)+p_t-p_r\right)
\right]\right).
\end{equation}
The relative rotation and translation map each target pixel to a location in the reference BEV, expressing the reference scene in the ego coordinate frame at time $t$.

During training, we construct the reprojected frames from an earlier annotated BEV frame and recorded ego poses, and supervise generation with the actual future BEV video. We VAE-encode the reprojected frames, add a small amount of noise ($\sigma_{\mathrm{cond}}=0.1$), and add their learned embeddings to the BEV input tokens through a zero-initialized projection. Joint attention then conveys this condition to the perspective branch. We use the same noise level during training and inference, and randomly drop the action condition during training to support generation without ego-action conditioning.

\subsubsection{Causal Rollout}
\label{subsec:method_rollout}

\begin{figure}[t]
  \centering
  \includegraphics[width=\linewidth]{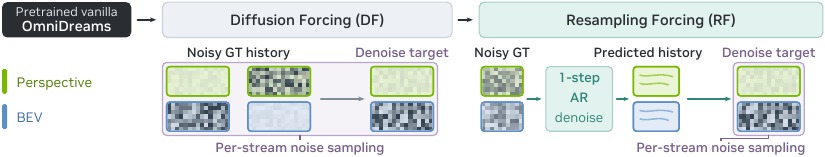}
  \caption{\textbf{Training procedure.} Starting from pretrained vanilla OmniDreams, we train with Diffusion Forcing (DF) using noise levels sampled independently across chunks and streams. During Resampling Forcing (RF) fine-tuning, noisy ground-truth chunks are autoregressively denoised with one Euler step per chunk to form the conditioning history. Both stages retain independent noise sampling across chunks and streams for target denoising.}
  \label{fig:training_pipeline}
\end{figure}

At inference, we first generate $B_0$ from $I_0$, then jointly generate the remaining perspective and BEV latents chunk by chunk while keeping the latent of $I_0$ fixed. Each subsequent chunk is denoised from Gaussian noise using cached keys and values from earlier generated chunks. When ego actions are provided, each chunk also receives the reprojected BEV condition described above. After generating a chunk, we cache its keys and values and update the BEV reference to its last frame. The next action condition then reflects the scene changes predicted in that chunk.

After DF training, we fine-tune with Resampling Forcing (RF)~\cite{resamplingforcing} to expose the model to errors in its own history (\cref{fig:training_pipeline}). RF corrupts the ground-truth chunks using Eq.~\eqref{eq:flow_interpolation} at a noise level $\tau$ sampled once per sequence and shared across both streams. It then constructs the history autoregressively by denoising each chunk with one Euler step.
\begin{equation}
\label{eq:resampled_history}
\widetilde z_k
= \operatorname{sg}\!\left[
x_k(\tau)-\tau\,v_\phi\!\left(x_k(\tau),\widetilde z_{<k},\tau\right)
\right],
\end{equation}
where $x_k(\tau)$ contains both corrupted streams, $\widetilde z_{<k}$ denotes the resampled history preceding the current chunk $k$, $v_\phi$ predicts their velocities, and $\operatorname{sg}$ stops gradients. Other conditioning inputs are omitted for brevity.

The shared $\tau$ is used only to construct the resampled history. For target denoising, we sample noise levels independently across chunks and streams and apply the flow-matching loss in Eq.~\eqref{eq:training_objective}. Each target chunk attends only to its own noisy tokens and the earlier resampled history $\widetilde z_{<k}$. All target chunks are trained in parallel.

\subsubsection{Flexible Conditioning}
\label{subsec:method_flexible}

Per-stream DF training also supports flexible conditioning at inference. We keep the latents of provided observations fixed and denoise the remaining latents from Gaussian noise, using the same trained model across input configurations. When a full video is provided in one stream, we generate the synchronized video in the other. An RGB video can therefore condition BEV video generation, while a BEV video can specify road geometry and vehicle motion for perspective video generation.

Conditioning can also be restricted to the first frame of either stream. Given an initial RGB frame $I_0$, we generate $I_{1:T}$ and $B_{0:T}$, including the initial BEV frame. Conversely, given an initial BEV frame $B_0$, we generate $I_{0:T}$ and $B_{1:T}$, including the initial perspective frame. An initial BEV layout can thus condition both the scene's appearance and its subsequent evolution. \Cref{fig:flexible-conditioning} illustrates joint generation from an initial RGB or BEV frame, as well as BEV video     generation from a full RGB video. Users can also edit vehicle positions and headings in $B_0$ before generating both videos, as illustrated by the BEV editing workflow in \cref{fig:bev-editing-tool}.

\subsection{Data Construction}
\label{subsec:method_data}

\noindent\textbf{Data sources.}
Each training example pairs a recorded front-camera RGB video with a schematic BEV video rendered from annotations of the same scene. We construct these pairs using a subset of RDS-HQ-1M~\cite{omnidreams} comprising approximately 670 hours of driving and the official nuScenes training split~\cite{nuscenes}. The nuScenes training set contains 700 scenes totaling approximately 3.9 hours of paired video. The RDS-HQ-1M subset provides road-layout annotations, tracked object boxes, ego poses, and camera calibration. We construct its road geometry from lanes, road boundaries, crosswalks, and wait lines. For nuScenes, we use its HD vector maps and annotated 3D box tracks.

\noindent\textbf{Track interpolation and filtering.}
For each RGB frame, we render a BEV frame using object positions and orientations at the time the RGB frame was captured. Box annotations may be less frequent than RGB frames. We therefore use the provided track IDs to identify annotations of the same object and interpolate its position and orientation between the earlier and later annotations. We also use these object tracks to interpolate vehicle positions and orientations during temporary occlusions and render the interpolated boxes in the BEV video. Before rendering, we remove duplicate or excessively overlapping vehicle boxes.

\noindent\textbf{Schematic BEV rendering.}
We render the road geometry and object boxes in an orthographic overhead view, with the ego vehicle at the center and its forward direction pointing upward. Road elements are drawn as polylines and polygons, and objects as oriented cuboids. Each BEV frame covers a square of approximately $69.3\,\mathrm{m}$ per side in both datasets, so the same pixel displacement represents the same physical distance across scenes. This pixel-to-meter mapping is also used for ego-action reprojection in Eq.~\eqref{eq:bev_reprojection}. Colors distinguish road elements and object categories: road boundaries are magenta, crosswalks purple, and pedestrians yellow. Vehicle cuboids are shaded red at the front and blue at the rear, making their headings visible. This representation preserves object positions and headings relative to the road without image texture or perspective foreshortening. The BEV video provides an explicit training target for vehicle motion relative to the road, complementing the appearance and motion supervision in the paired RGB video.

\begin{table}[t]
  \centering
  \caption{\textbf{Quantitative comparison without ego-action conditioning.} Each entry reports PhysicalAI-AV~/~nuScenes. Trajectory metrics use ego motion reconstructed from generated perspective videos; minADE and minDTW report the minimum over three samples per scene against the logged trajectory. $^*$OmniDreams (Vanilla) uses no external scene-layout conditioning and is post-trained on the same dataset as \ours.}
  \label{tab:main-uncond}
  \resizebox{\linewidth}{!}{%
  \begin{tabular}{l ccc ccc}
    \toprule
    & \multicolumn{3}{c}{\emph{Ego-Trajectory Metrics}} & \multicolumn{3}{c}{\emph{Perspective Video Quality}} \\
    \cmidrule(lr){2-4} \cmidrule(lr){5-7}
    Method & minADE~$\downarrow$ & minDTW~$\downarrow$ & FTD~$\downarrow$ & FVD~$\downarrow$ & FID~$\downarrow$ & Image Qual.~$\uparrow$ \\
    \midrule
    Vista & 14.65~/~8.70 & 890~/~512 & 9.02~/~14.24 & 388~/~460 & 27.34~/~30.75 & 0.363~/~0.360 \\
    UniFuture & 17.53~/~7.16 & 1069~/~432 & 18.11~/~10.34 & 535~/~393 & 35.50~/~30.47 & 0.363~/~0.366 \\
    GEM & 10.43~/~7.03 & 580~/~420 & 4.00~/~12.42 & 359~/~401 & 23.45~/~29.84 & 0.385~/~0.388 \\
    DrivingDojo & 14.81~/~11.43 & 967~/~744 & 14.58~/~15.44 & 367~/~469 & 32.56~/~34.21 & 0.395~/~\rankfirst{0.443} \\
    \midrule
    OmniDreams (Vanilla)$^*$ & \ranksecond{7.76}~/~\ranksecond{6.20} & \ranksecond{413}~/~\ranksecond{393} & \ranksecond{3.17}~/~\ranksecond{4.15} & \ranksecond{281}~/~\ranksecond{351} & \ranksecond{11.37}~/~\ranksecond{15.58} & \rankfirst{0.445}~/~0.430 \\
    \ours & \rankfirst{6.15}~/~\rankfirst{3.91} & \rankfirst{258}~/~\rankfirst{205} & \rankfirst{1.77}~/~\rankfirst{0.43} & \rankfirst{271}~/~\rankfirst{309} & \rankfirst{10.70}~/~\rankfirst{11.48} & \ranksecond{0.428}~/~\ranksecond{0.439} \\
    \bottomrule
  \end{tabular}%
  }
\end{table}

\begin{table}[t]
  \centering
  \caption{\textbf{Quantitative comparison with ego-action conditioning.} Each entry reports PhysicalAI-AV~/~nuScenes.  OmniDreams (Pl\"ucker) conditions perspective video on ego motion through Pl\"ucker ray embeddings.}
  \label{tab:main-cond}
  \resizebox{\linewidth}{!}{%
  \begin{tabular}{l ccc ccc}
    \toprule
    & \multicolumn{3}{c}{\emph{Ego-Trajectory Metrics}} & \multicolumn{3}{c}{\emph{Perspective Video Quality}} \\
    \cmidrule(lr){2-4} \cmidrule(lr){5-7}
    Method & ADE~$\downarrow$ & DTW~$\downarrow$ & FTD~$\downarrow$ & FVD~$\downarrow$ & FID~$\downarrow$ & Image Qual.~$\uparrow$ \\
    \midrule
    Vista & 9.65~/~6.65 & 343.6~/~298.9 & 4.75~/~12.11 & 341.2~/~407.8 & 21.23~/~25.89 & 0.396~/~0.405 \\
    UniFuture & 10.33~/~4.28 & 361.2~/~\ranksecond{175.3} & 7.87~/~10.08 & 608.0~/~349.2 & 44.63~/~33.11 & 0.361~/~0.371 \\
    GEM & 6.59~/~5.55 & 235.6~/~245.0 & \ranksecond{1.99}~/~9.06 & 336.1~/~386.1 & 24.50~/~32.29 & 0.394~/~0.406 \\
    DrivingDojo & 9.87~/~\ranksecond{3.81} & 390.1~/~225.2 & 9.21~/~8.26 & 410.3~/~406.8 & 27.28~/~40.61 & 0.404~/~0.411 \\
    \midrule
    OmniDreams (Pl\"ucker) & \ranksecond{5.88}~/~5.38 & \ranksecond{197.1}~/~234.8 & 3.02~/~\ranksecond{1.47} & \ranksecond{271.5}~/~\ranksecond{343.1} & \ranksecond{15.35}~/~\ranksecond{18.27} & \ranksecond{0.416}~/~\rankfirst{0.443} \\
    \ours & \rankfirst{3.83}~/~\rankfirst{3.42} & \rankfirst{95.2}~/~\rankfirst{133.3} & \rankfirst{0.48}~/~\rankfirst{0.56} & \rankfirst{238.9}~/~\rankfirst{294.0} & \rankfirst{11.81}~/~\rankfirst{14.80} & \rankfirst{0.435}~/~\ranksecond{0.436} \\
    \bottomrule
  \end{tabular}%
  }
\end{table}

\begin{figure}[t]
  \centering
  \includegraphics[width=\linewidth]{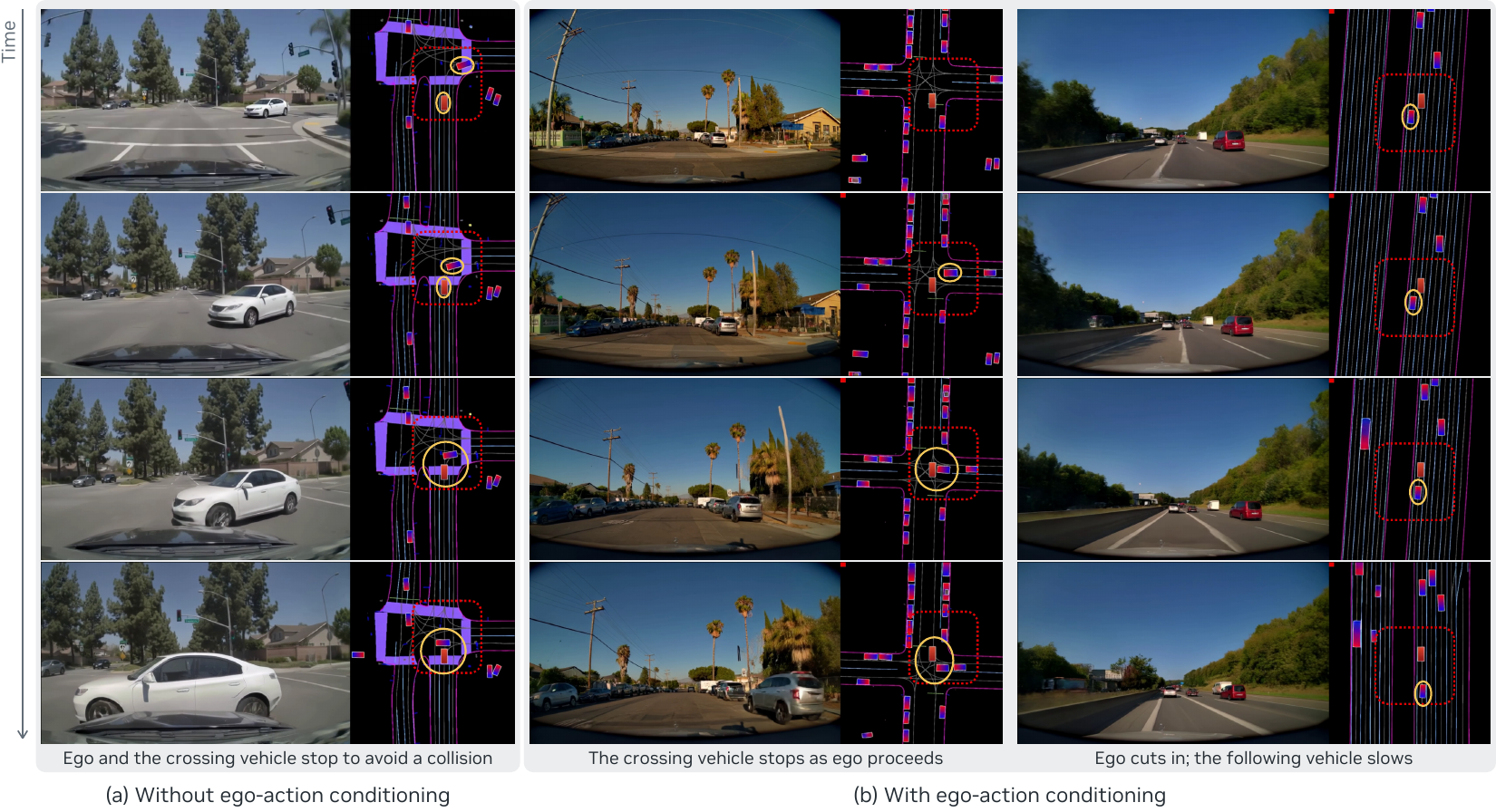}
  \caption{\textbf{Reactive behavior with and without ego-action conditioning.} (a)~Without ego-action conditioning, the ego and surrounding vehicles respond to one another. (b)~In both examples with ego-action conditioning, the ego follows the input trajectory while surrounding vehicles react to its motion.}
  \label{fig:counterfactual}
\end{figure}
\begin{figure}[p]
  \centering
  \includegraphics[width=\linewidth,height=0.87\textheight,keepaspectratio]{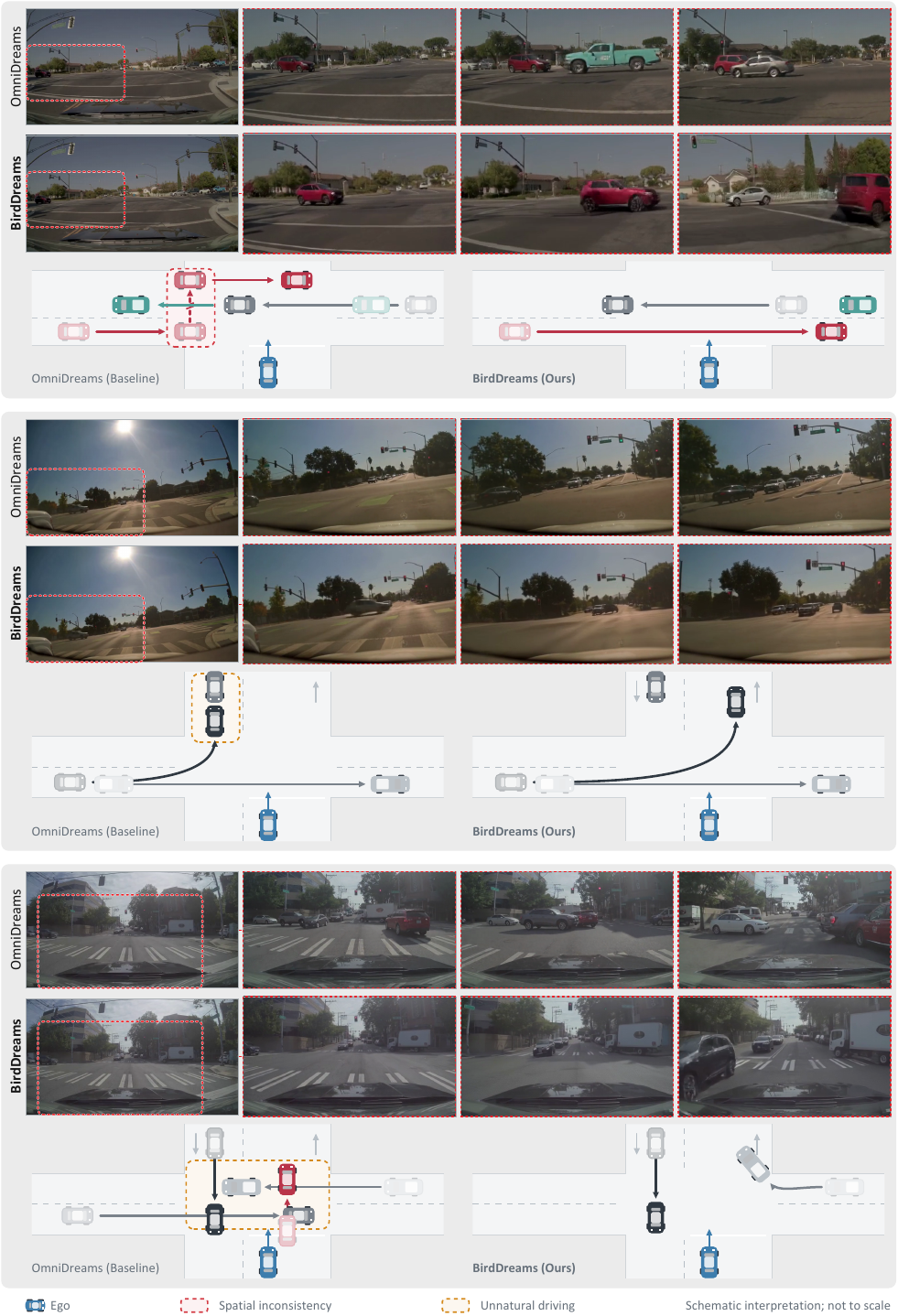}
  \caption{\textbf{Surrounding-vehicle behavior under counterfactual ego actions.} We compare HD-map-conditioned variants of OmniDreams and \ours using the same initial observations, without surrounding-vehicle-state conditions. The schematics summarize vehicle motion, with faded-to-solid symbols indicating earlier-to-later positions and solid arrows showing motion directions. Compared with the OmniDreams baseline, \ours generates more realistic driving maneuvers while maintaining spatially coherent vehicle motion over time.}
  \label{fig:counterfactual-baselines}
\end{figure}

\section{Experiments}
\label{sec:experiments}

We evaluate perspective video quality and ego-trajectory control with and without ego-action conditioning, then examine surrounding-vehicle responses to counterfactual ego maneuvers through qualitative examples and human evaluation. Additional experiments examine how independent noise sampling affects generation quality, explore ego-trajectory prediction through post-training, and demonstrate scenario editing in BEV.

\subsection{Experimental Setup}
\label{sec:exp-setup}

\noindent\textbf{Implementation details.}
\ours is trained from a checkpoint of an OmniDreams variant without external scene-layout conditioning (HD maps and surrounding-vehicle states), on a subset of RDS-HQ-1M~\cite{omnidreams} comprising approximately 670 hours of driving. Perspective frames ($848\times480$) and schematic BEV frames ($480\times480$) are encoded by a causal video VAE and modeled by a ${\sim}4$B dual-branch Mixture-of-Transformers diffusion transformer with block-causal attention.
Training proceeds in two stages: the model is first trained with Diffusion Forcing~\cite{diffusionforcing}, and then further trained with Resampling Forcing~\cite{resamplingforcing} to reduce error accumulation over autoregressive rollouts. We train on 64 H100 GPUs with a total batch size of 64, using AdamW with a learning rate of $3\times10^{-5}$, weight decay of $10^{-3}$, and a linear learning-rate schedule with 1,000 warm-up steps.

\noindent\textbf{Benchmarks.}
We evaluate on two benchmarks. \emph{PhysicalAI-AV} is a subset we construct from PhysicalAI-Autonomous-Vehicles~\cite{physicalaiav}, a public driving video dataset released by NVIDIA, by randomly sampling 200 clips from its evaluation split. \emph{nuScenes}~\cite{nuscenes} is a public driving video dataset that enables comparison in a widely used setting; for this comparison, we further post-train our model on the nuScenes train split and evaluate on all 150 scenes in its official validation split.

\noindent\textbf{Baselines.}
We compare against open driving world models: Vista~\cite{gao2024vista}, UniFuture~\cite{unifuture}, GEM~\cite{gem}, and DrivingDojo~\cite{drivingdojo}, evaluating each with and without ego-action conditioning. Without ego-action conditioning, we additionally compare against OmniDreams (Vanilla), a variant of the OmniDreams series without external scene-layout conditioning (HD maps and surrounding-vehicle states), post-trained on the same dataset as \ours. With ego-action conditioning, Vista, UniFuture, and GEM receive ego waypoints, while DrivingDojo receives per-frame ego displacements. The OmniDreams baseline for this setting is OmniDreams (Pl\"ucker), which conditions on ego motion through Pl\"ucker ray embeddings.

\noindent\textbf{Evaluation settings and metrics.}
Each benchmark is evaluated in two settings: \emph{without ego-action conditioning}, where the model rolls out from the conditioning frame(s) alone, and \emph{with ego-action conditioning}, where the ground-truth ego trajectory is additionally provided as the action condition. Our metrics largely follow the DrivingGen protocol~\cite{drivinggen} and cover two axes. For perspective video quality, we report the Fr\'echet Video Distance (FVD)~\cite{fvd}, the Fr\'echet Inception Distance (FID)~\cite{fid}, and Image Quality (reference-free perceptual quality via CLIP-IQA+~\cite{clipiqa}). For ego-trajectory quality, we report the Fr\'echet Trajectory Distance (FTD), computed between ego-trajectory embeddings from a pretrained motion-forecasting encoder~\cite{mtr}. To compare ego motion with models that generate only perspective video, we reconstruct ego trajectories from generated and corresponding real perspective videos using Depth Anything~3~\cite{da3}. We measure agreement with the trajectories from real videos using average displacement error (ADE) and dynamic time warping (DTW). ADE averages the Euclidean distance between their 2D positions at corresponding timestamps. DTW computes the minimum sum of pairwise Euclidean distances over monotonic temporal alignments. We find this alignment by dynamic programming, allowing one-to-many position matches to accommodate differences in traversal speed~\cite{drivinggen}. With ego-action conditioning, we evaluate one generated sample per scene. Without ego-action conditioning, where many futures are plausible, we generate three samples per scene with different random seeds and take the minimum ADE and DTW independently for each scene, denoted minADE and minDTW.

\subsection{Video Generation and Ego-Trajectory Control}
\label{sec:exp-comparison}

We compare \ours with existing open driving world models on PhysicalAI-AV and nuScenes, without ego-action conditioning (\cref{tab:main-uncond}) and with ego-action conditioning (\cref{tab:main-cond}). Without ego-action conditioning, every model receives only the conditioning frames, so the metrics jointly assess the visual fidelity of the generated perspective video and the plausibility of the generated ego motion. With ego-action conditioning, the ground-truth ego trajectory is additionally provided, and ADE and DTW quantify how faithfully each model executes it.

\noindent\textbf{Results.}
Without ego-action conditioning, \ours achieves the lowest minADE, minDTW, FTD, FVD, and FID among the evaluated methods on both benchmarks (\cref{tab:main-uncond}). It reduces minADE relative to OmniDreams (Vanilla) by 21\% on PhysicalAI-AV and 37\% on nuScenes, indicating closer agreement with logged ego behavior among the sampled futures. For frame-level Image Quality, \ours ranks second to OmniDreams (Vanilla) on PhysicalAI-AV and DrivingDojo on nuScenes.

With ego-action conditioning, \ours follows the input ego trajectory more accurately in the generated perspective videos (\cref{tab:main-cond}), reducing ADE by 35\% on PhysicalAI-AV and 36\% on nuScenes relative to OmniDreams (Pl\"ucker). Trajectory improvements in both settings are accompanied by lower FVD and FID. For per-frame Image Quality, \ours ranks first on PhysicalAI-AV and second on nuScenes. These benchmarks use logged ego trajectories as action conditions; we next examine surrounding-vehicle behavior when the ego departs from those trajectories.

\subsection{Reactive Behavior of Surrounding Vehicles}
\label{sec:exp-behavior}
\label{sec:exp-counterfactual}

\noindent\textbf{Counterfactual scenarios.}
A world model useful for closed-loop simulation must generate videos in which surrounding vehicles \emph{react} to the ego vehicle, rather than replaying logged motion. We probe this with counterfactual rollouts: starting from the same conditioning frames, we provide ego actions that differ from the ground-truth log and observe how the vehicles visible in the first frame behave. \Cref{fig:counterfactual} shows selected vehicle interactions with and without ego-action conditioning. Without ego-action conditioning, the ego and a crossing vehicle stop. With ego-action conditioning, a crossing vehicle stops as the ego proceeds, and a following vehicle slows after the ego cuts into its lane. In extreme scenarios that leave no room to react, the rollout ends in a collision between the ego and another vehicle, which can be a plausible outcome under such maneuvers.

\noindent\textbf{Comparison with baselines.}
To compare surrounding-vehicle behavior with and without vehicle-motion conditioning, we consider two OmniDreams-based baselines. The original OmniDreams conditions on surrounding-vehicle motion replayed from driving logs, so the depicted vehicles follow recorded motion under counterfactual ego actions. To focus the evaluation on the behavior of non-ego agents, we compare both models under the same static road geometry. We train OmniDreams (HDMap) with surrounding vehicles removed from its perspective layout conditions, retaining static road geometry. For \ours (HDMap), we add the same road geometry as a BEV condition, since the standard model may generate roads that differ from the map in areas not visible in the initial image. Conditioned on the surrounding road, terrain, and ego actions, OmniDreams (HDMap) must predict surrounding-vehicle behavior through perspective video alone. As shown in \cref{fig:counterfactual-baselines}, its rollouts exhibit failures including spatially inconsistent motion and unnatural driving. The red car appears to switch sides relative to oncoming traffic, while vehicles in the intersection example proceed along conflicting paths without a coherent interaction. In these examples, \ours generates more plausible vehicle motion under the same conditions.

\begin{figure}[p]
  \centering
  \includegraphics[width=\linewidth,height=0.87\textheight,keepaspectratio]{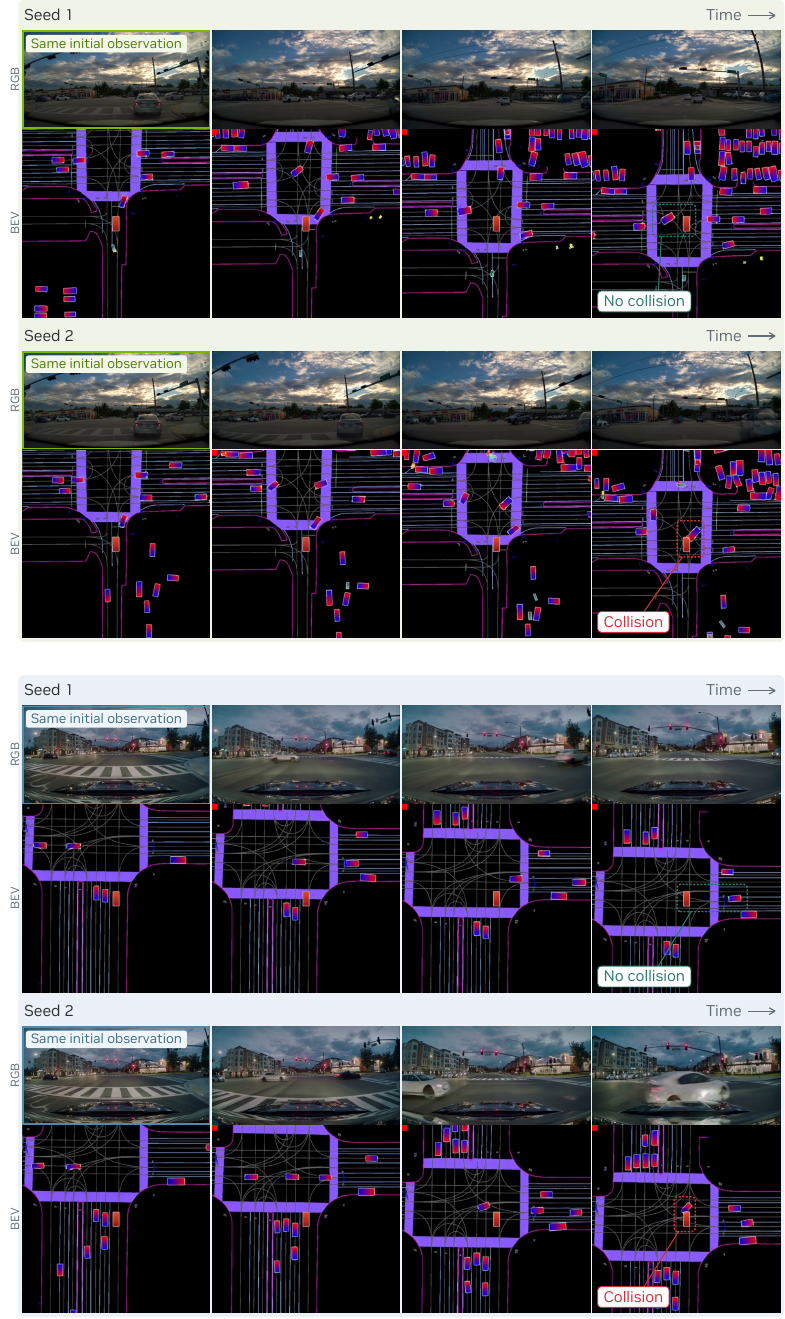}
  \caption{\textbf{Diverse futures from the same initial observation and ego action.} We show RGB and BEV rollouts from \ours (HDMap) to examine surrounding-vehicle responses to counterfactual ego actions that may lead to collisions. Each pair uses the same initial RGB observation, counterfactual ego trajectory, and static road geometry, but different random seeds. Different surrounding-vehicle motions yield a collision in one rollout and no collision in the other.}
  \label{fig:multiseed-reactive}
\end{figure}

\noindent\textbf{Human evaluation.}
Since the realism of reactive behavior is difficult to capture with automatic metrics, we conduct a pairwise human evaluation. We generate paired rollouts of \ours (HDMap) and OmniDreams (HDMap) from the same conditioning frames, counterfactual ego actions, and static road geometry, and show each of 25 participants 15 pairs randomly sampled from a pool of 119 pairs in randomized order, yielding $N=375$ pairwise judgments. For each pair, participants answer in which video the surrounding vehicles move and behave more realistically (Video A, Video B, or about the same), with the instruction that a dangerous situation or a crash is a realistic outcome when the ego maneuver leaves no time to avoid it. When a participant prefers one video, they additionally flag everything wrong with the other one, choosing from impossible motion (teleporting, sliding, passing through objects, or sudden appearance/disappearance), unnatural driving (driving against traffic or wrong spacing), no appropriate reaction to the ego or other vehicles (e.g., no braking or yielding when another car cuts in), and other.

\begin{figure}[t]
  \centering
  \includegraphics[width=0.95\linewidth]{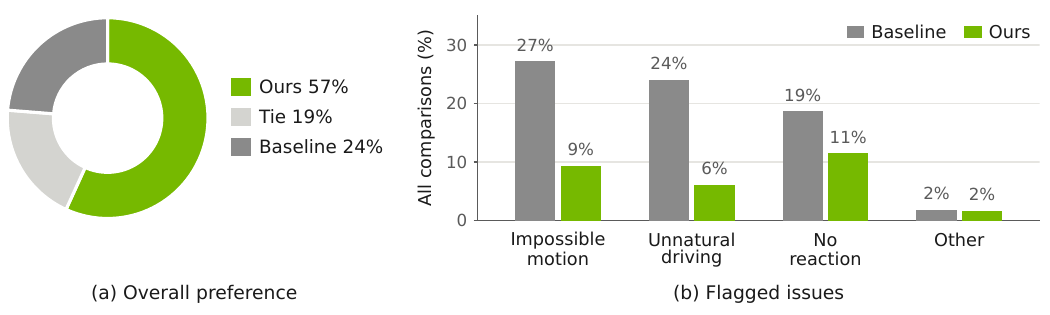}
  \caption{\textbf{Human evaluation of surrounding-vehicle behavior.} \ours (HDMap) and OmniDreams (HDMap) receive the same initial observations, counterfactual ego actions, and static road geometry, without surrounding-vehicle-state conditions. (a)~Overall preference from 375 judgments by 25 participants. (b)~Issues flagged for the non-preferred video, with percentages over all 375 judgments and multiple selections allowed.}
  \label{fig:user-study}
\end{figure}

Participants prefer \ours (HDMap) in 57\% of comparisons, versus 24\% for OmniDreams (HDMap), with 19\% ties (\cref{fig:user-study}a). This preference supports more realistic surrounding-vehicle behavior under the tested counterfactual ego maneuvers, with road geometry given and vehicle motion left for the models to predict. The flagged issues indicate where the methods differ (\cref{fig:user-study}b): the baseline is flagged for impossible motion in 27\% of comparisons, for unnatural driving in 24\%, and for missing reactions in 19\%, compared with 9\%, 6\%, and 11\% for \ours (HDMap). Overall, participants prefer \ours (HDMap) for the realism of surrounding-vehicle behavior under the same counterfactual ego actions.

\noindent\textbf{Diverse future scenarios.}
Surrounding vehicles can respond differently to the same ego maneuver, so a world model should be able to simulate multiple possible futures from a given initial scene. We examine this capability by varying the random seed in \ours (HDMap) while fixing the initial RGB observation, counterfactual ego trajectory, and static road geometry. The sampled rollouts in \cref{fig:multiseed-reactive} exhibit distinct sequences of vehicle interactions, illustrating how \ours can simulate different ways the same scene may unfold under a fixed ego action.

\subsection{Model Analysis and Extensions}
\label{sec:exp-analysis}

\subsubsection{Per-Stream Noise Sampling}
\label{sec:exp-ablation}

Independent noise sampling exposes the model to cases in which one stream is cleaner than the other, allowing each branch to learn denoising with different levels of information from the complementary stream (\cref{subsec:method_df}). We compare it with a DF variant that shares one noise level across the perspective and BEV streams within each chunk. Both variants sample noise levels independently across temporal chunks. \Cref{fig:per-stream-noise} reports FVD and FTD for the DF variant on RDS-HQ-1M validation clips. Independent sampling yields lower FVD and FTD than shared sampling.

\subsubsection{World-Action Model Post-Training}
\label{sec:exp-wam}

We conduct a preliminary experiment to explore extending \ours to a world-action model (WAM) for ego-trajectory prediction. We add an ego-action prediction head to its perspective backbone and apply the same post-training procedure to OmniDreams (Vanilla) to examine whether joint perspective--BEV video modeling benefits ego-trajectory prediction. Both models are post-trained for 22,000 iterations on our RDS-HQ-1M subset, fine-tuning the perspective branch and action head. \ours retains BEV supervision with its BEV branch frozen.

\begin{figure}[t]
  \centering
  \begin{minipage}[t]{0.39\linewidth}
    \vspace{0pt}
    \centering
    \includegraphics[width=\linewidth]{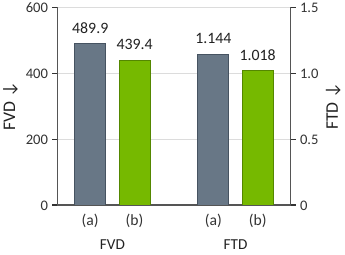}
    \caption{\textbf{Per-stream noise sampling.} (a) Shared noise levels across perspective and BEV streams; (b) independent noise levels (\ours).}
    \label{fig:per-stream-noise}
  \end{minipage}\hfill
  \begin{minipage}[t]{0.57\linewidth}
    \vspace{0pt}
    \centering
    \includegraphics[width=\linewidth]{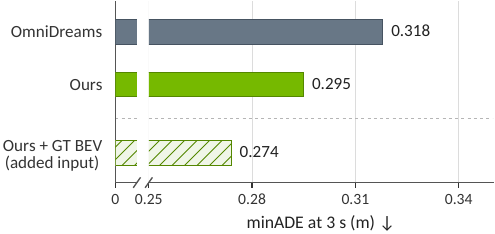}
    \caption{\textbf{Ego-trajectory prediction after WAM post-training.} Best-of-six minADE over 3\,s on 100 held-out RDS-HQ-1M clips. GT BEV (hatched) is a separate diagnostic with additional ground-truth BEV input.}
    \label{fig:wam}
  \end{minipage}
\end{figure}

We evaluate on 100 held-out clips from this subset. Both models use perspective video as their only visual observation, together with 1.6 seconds of ego-motion history. We therefore feed pure Gaussian noise ($\sigma=1$) to the BEV branch. The branch participates in joint attention, and its input is not iteratively denoised during action prediction. We report best-of-six minADE over a three-second horizon, averaged across clips. As a separate diagnostic, we additionally condition the model on ground-truth BEV across all input video frames at a low noise level (GT BEV). This diagnostic uses privileged observations beyond the perspective video. As shown in \cref{fig:wam}, \ours achieves 0.295\,m minADE, compared with 0.318\,m for OmniDreams (Vanilla), a numerical improvement of 7.3\%. The GT BEV diagnostic reduces the error to 0.274\,m.

\begin{figure}[!t]
  \centering
  \includegraphics[width=\linewidth]{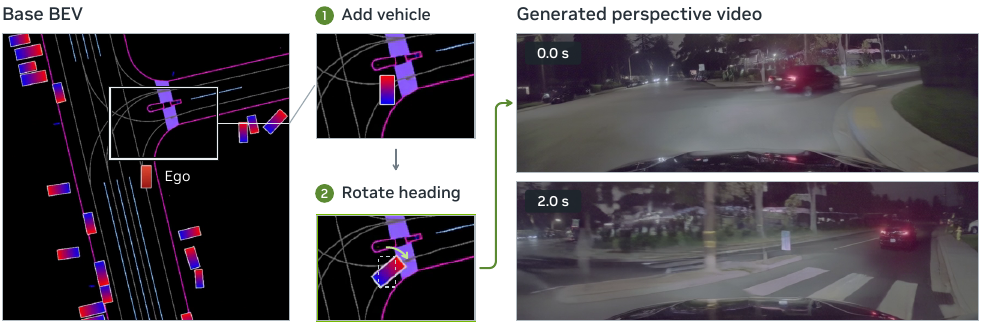}
  \caption{\textbf{Scenario editing in BEV.} The schematic top view makes it easy to customize driving scenarios by adding vehicles and adjusting their positions and headings. \ours then jointly generates perspective and BEV videos from this initial layout, synthesizing scene appearance and subsequent vehicle motion without requiring a corresponding perspective image.}
  \label{fig:bev-editing-tool}
\end{figure}

\subsubsection{Scenario Editing in BEV}
\label{sec:exp-editing}

Flexible conditioning (\cref{subsec:method_flexible}) lets users customize driving scenarios by editing an initial BEV frame. Starting from an existing driving log, users can add vehicles or adjust their positions and headings in our editor, as illustrated in \cref{fig:bev-editing-tool}. These edits are made directly in the projected top view, avoiding manipulation of 3D bounding boxes. Given the edited BEV, \ours jointly generates perspective and BEV videos, including the first perspective frame, so users need not provide a perspective image of the modified scene.

In the example shown in \cref{fig:bev-editing-tool}, the edited BEV places a vehicle ahead of the ego vehicle at an intersection. From this initial layout, \ours generates a rollout in which the vehicle turns right onto the intersecting road and the ego vehicle follows.

\section{Conclusion}
\label{sec:conclusion}

We have presented \ours, an autoregressive driving world model that jointly generates perspective and schematic BEV videos. Our model learns road-relative vehicle motion from BEV supervision and uses the predicted BEV to condition generation on an ego trajectory. On PhysicalAI-AV and nuScenes, \ours achieves lower ego-trajectory errors than prior methods with and without ego-action conditioning. Qualitative counterfactual rollouts show surrounding vehicles reacting to changed ego maneuvers, such as a following vehicle slowing after an ego cut-in. In a human evaluation where both models received the same road geometry, participants also preferred its surrounding-vehicle behavior to that of a perspective-only baseline. These results suggest that predicting scene geometry alongside appearance can improve ego-trajectory control and the realism of surrounding-vehicle responses in driving world models.

\section*{Acknowledgments}
We are deeply grateful to Sanja Fidler, Zan Gojcic, Xuanchi Ren, Kevin Xie, Yifan Lu, and Zian Wang for their valuable discussions, insightful feedback, and generous support throughout this work. We also thank Yuxuan Zhang for developing the HD-map-conditioned variant of OmniDreams.

\clearpage
{
\small
\bibliographystyle{abbrv}
\bibliography{bibs/reference,bibs/reference2}
}
\end{document}